# Shaping Individualized Impedance Landscapes for Gait Training via Reinforcement Learning

Yufeng Zhang, *Student Member, IEEE*, Shuai Li, *Student Member, IEEE,* Karen J. Nolan, and Damiano Zanotto, *Member, IEEE*

*Abstract*—Assist-as-needed (AAN) control aims at promoting therapeutic outcomes in robot-assisted rehabilitation by encouraging patients' active participation. Impedance control is used by most AAN controllers to create a compliant force field around a target motion to ensure tracking accuracy while allowing moderate kinematic errors. However, since the parameters governing the shape of the force field are often tuned manually or adapted online based on simplistic assumptions about subjects' learning abilities, the effectiveness of conventional AAN controllers may be limited. In this work, we propose a novel adaptive AAN controller that is capable of autonomously reshaping the force field in a phase-dependent manner according to each individual's motor abilities and task requirements. The proposed controller consists of a modified *Policy Improvement with Path Integral* algorithm, a model-free, sampling-based reinforcement learning method that learns a subject-specific impedance landscape in real-time, and a hierarchical policy parameter evaluation structure that embeds the AAN paradigm by specifying performance-driven learning goals. The adaptability of the proposed control strategy to subjects' motor responses and its ability to promote short-term motor adaptations are experimentally validated through treadmill training sessions with able-bodied subjects who learned altered gait patterns with the assistance of a powered ankle-foot orthosis.

## I. INTRODUCTION

IN motor rehabilitation, patients who suffered a neurological injury such as stroke, spinal cord injury, or traumatic brain injury must re-learn the correct spatiotemporal muscle activation pattern to achieve a target movement through exercise therapy. The ultimate goal of rehabilitation robotics is to encourage the recovery of motor ability in these patients by automating exercise protocols that are traditionally administered by physical therapists [1]. Compared with manual assistance, robot-assisted training has the potential to increase frequency and intensity of treatments and enable highly repetitive practice, all of which have been shown to benefit rehabilitation outcomes [2]. However, despite over two decades of research into robot-assisted rehabilitation, how to control a robot to best promote motor recovery is still an open research problem [3].

The study of how able-bodied individuals adapt to robot-applied forces while performing a motion task has been used as a paradigm to understand mechanisms of motor re-learning after neurological injury, since motor adaptation and motor recovery share similar learning processes [4]. Pioneering studies on perturbed reaching [5] and walking tasks [6] indicated that the human motor control system progressively adapts to robot-applied forces by forming internal feed-forward models. These models have been mathematically interpreted as P-type iterative learning control (ILC) rules with forgetting term, which greedily optimize the weighted sum of movement error and physical effort at each movement repetition [6]. As a result, during robot-assisted training, individuals may let the robot "take over" the motion tasks completely – if given the opportunity – thereby reducing their physical effort [7]. Because active effort is a critical enabler of motor learning, this "slacking" effect is detrimental for robot-assisted training [8]–[10]. To mitigate the slacking effect, rehabilitation robots should provide only the amount of assistance required for a subject to complete the target motion and should allow a controlled level of kinematic error to encourage learning. Controllers designed to achieve these goals are called *assist-as-needed* (AAN) controllers [11], [12].

Early AAN controllers aimed at adapting the level of robotic assistance by following an ILC law that approximates the way the human motor control system adapts to a training task [13]. To mitigate the "slacking effect", these ILC-AAN controllers must reduce the assistive forces at a faster rate than the rate at which subjects reduce their effort [14]. Later studies on ILC-AAN replaced the explicit human learning model with heuristic parameters, thereby expanding its applicability at the expense of introducing time-consuming tuning processes [15]. More recent implementations of the AAN paradigm rely on adaptive error-dependent force fields shaped by an underlying impedance control law [16], [17]. The force field provides zero assistance in the proximity of the target trajectory – thereby allowing small kinematic errors – while establishing a relationship between kinematic errors and corrective forces to assist the subject when movements deviate substantially from the target trajectory. These AAN controllers require less tuning than ILC-AAN controllers, however they do not self-adapt to an individual's motor ability.

Reinforcement Learning (RL) solves optimal control problems by learning a control policy online, through repeated interactions between a learning agent and its environment [18]. Applications of RL in wearable robotics have so far focused on self-tuning of assistive exoskeletons [19]–[23] and active prostheses [24]–[26], with only a handful of studies addressing robot-assisted rehabilitation [27]–[29]. In assistive

Y. Zhang, S. Li and D. Zanotto (dzanotto@stevens.edu) are with the Wearable Robotic Systems Lab., Department of Mechanical Engineering, Stevens Institute of Technology, Hoboken, NJ, 07030 USA.
K. J. Nolan is with Human Performance and Engineering Research, Kessler Foundation, West Orange, NJ, USA and Rutgers-NJMS, Newark, NJ, USA.

exoskeletons and active prostheses, the goal of the RL-based controller is to track a predefined trajectory or minimize human physical effort. However, in rehabilitation robots, the goal is profoundly different, since tracking errors and human efforts are enabling factors rather than costs [30]. These two factors should be continuously balanced (in a subject-specific fashion) during the human learning process by the adaptive controller, until the robot's assistance is no longer needed for the subject to complete the target motion task, in line with the AAN paradigm. Because the goal-oriented behavior and the reward-based feedback of a RL agent echo the way humans learn new motor skills [31], we argue that the RL framework holds great potential for the design of more individualized adaptive controllers in rehabilitation robotics.

Based on action-dependent heuristic dynamic programming (ADHDP) [32], our previous work on robot-assisted gait training (RAGT) incorporated the AAN paradigm into the RL framework via the actor-critic method [29], [33]. By correlating the actor control objective with the subject's tracking errors in recent gait cycles, the ADHDP-AAN control framework automatically stiffens the assistive force field when the subject struggles to complete the training task, and progressively increases compliance when sustained good performances are detected. Unlike traditional ILC-AAN strategies that diminish assistance levels at a predefined rate [14], the ADHDP-AAN approach does not rely on presumptions about one's learning ability and can modify the control law in real-time based on an individual's motor responses. However, the adaptation of the ADHDP-AAN controller was limited to the stride level (i.e., the same stiffness parameter holds in the entire gait cycle). Because gait abnormalities may vary substantially between different individuals and may affect certain gait phases more than the others [34], [35], *a more desirable control strategy should adapt the control law at the gait phase level*.

In this work, we introduce a modified *Policy Improvement with Path Integral* ($PI^2$) algorithm which greatly extends the adaptability of the ADHDP-AAN control strategy described above, by enabling individualized and phase-dependent assistance. The impedance of the force field is parameterized by a set of phase-locked Gaussian-like kernel functions that are equally spaced across the gait cycle. An impedance landscape is formed by the weighted sum of the kernel functions. The underlying weight parameters are learned and updated through a $PI^2$ algorithm adapted from the work of Theodorou et al. [36]. The uniqueness of the proposed approach lies in the integration of the AAN paradigm into the $PI^2$ framework, which is achieved through a hierarchical policy parameter evaluation structure that specifies different learning objectives based on the subject's training progress. Specifically, the high-level evaluation is directly linked to the subject's recent training performances and is used to determine the learning modes for the low-level policy iterations, to raise or recede the impedance landscape as needed. In contrast to the conventional $PI^2$ algorithm that terminates when the trajectory cost falls within a certain limit, the proposed hierarchical evaluation approach allows the $PI^2$-AAN algorithm to operate continuously during gait training sessions while mimicking the behavior of a physical therapist who constantly modifies the assistance levels with the goal of promoting motor learning.

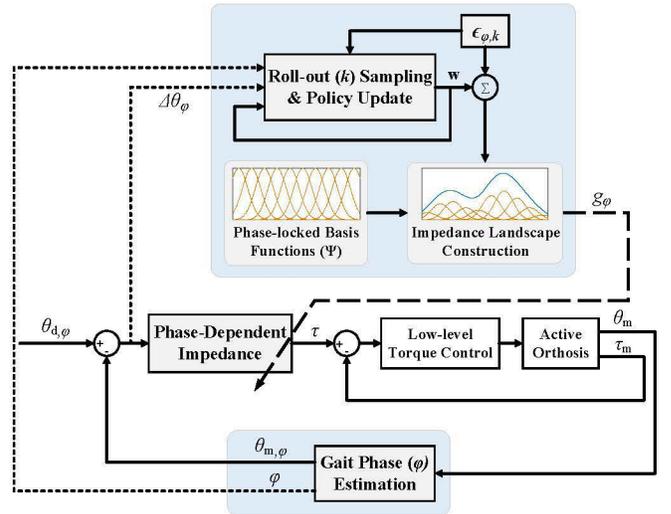

Fig. 1. Control scheme of the $PI^2$-AAN controller. The $PI^2$ procedure for impedance landscape modulation is shown in the upper shaded area ($\varphi$: gait phase; $k$ stride index; $g$ impedance coefficient).

This paper is organized as follows. Sec. II introduces the $PI^2$-AAN controller. In Sec. III, the adaptability of the $PI^2$-AAN controller and its effectiveness in promoting human short-term motor adaptation is evaluated through walking experiments wherein able-bodied individuals were asked to learn a target gait pattern with the help of a powered ankle-foot orthosis. Experimental results are presented in Sec. IV and discussed in Sec. V. We conclude with additional remarks in Sec. VI.

## II. AAN Control with $PI^2$

Figure 1 illustrates the block diagram of the proposed $PI^2$-AAN controller for RAGT. We define the impedance control law that shapes the underlying phase-dependent assistive force field as

$$|\tau| = \tau_{\max}[1 - e^{-(\Delta\theta\, g)^2}], \quad (1)$$

where $\Delta\theta$ indicates the tracking error between the desired trajectory $\theta_d$ and the measured trajectory $\theta_m$, and $g$ dictates the impedance of the assistive force field. As we shall discuss in Sec. II-A, $g$ is a function of the current gait phase $\varphi$. $\tau_{\max}$ determines the upper bound in the magnitude of the assistive torque $\tau$. The direction of $\tau$ is determined by $\text{sign}(\Delta\theta)$, such that the force field exerts a restoring action to assist the limb motion towards the desired path. A deadband $\theta_{\text{db}}$ is introduced to impose null assistance in the proximity of $\theta_d$, so that slight gait variations can be properly accommodated [37]. The relationship between $\theta_d$, $\theta_m$, $\theta_{\text{db}}$, and $\Delta\theta$ is expressed as follows:

$$\Delta\tilde{\theta} = \theta_d - \theta_m \quad (2)$$

$$\Delta\theta = \begin{cases} \text{sign}(\Delta\tilde{\theta})\left(|\Delta\tilde{\theta}| - \theta_{\text{db}}\right) & \text{if } |\Delta\tilde{\theta}| \geq \theta_{\text{db}}, \\ 0 & \text{if } |\Delta\tilde{\theta}| < \theta_{\text{db}}. \end{cases} \quad (3)$$

The key feature that sets the $PI^2$-AAN apart from conventional AAN controllers [16], [38]–[40] and from our previous work on RL controllers [29], [33] is its ability to continuously modulate the impedance coefficient $g$ in a phase-dependent manner based on the subject's training performance. This adaptability is made possible through the parameterization of $g$ using phase-locked shape parameters. In the following subsections, we describe the phase-locked Gaussian basis functions that generate the impedance landscape $g(\varphi)$, the online adaptation of the impedance landscape through a modified $PI^2$ algorithm, and the dual-objective policy evaluation rule that embeds the AAN paradigm in the $PI^2$ algorithm by updating the learning goal based on the wearer's recent training performance.

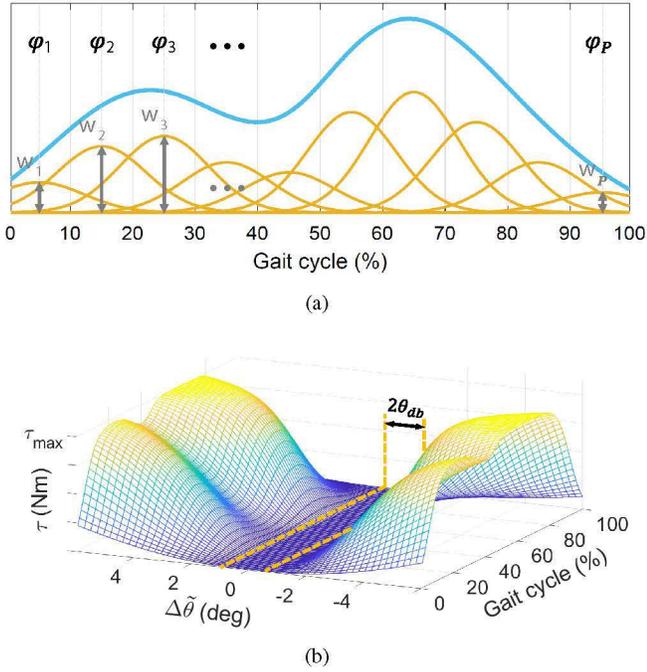

Fig. 2. (a) The set of phase-locked Gaussian basis functions $\Psi(\varphi)$ centered at $\varphi_{1 \sim P}$ (orange curves) and the associated set of adaptive shape parameters $\mathbf{w}$ determine the impedance landscape $g(\varphi)$ (blue curve) according to (4); (b) Assistive force field generated from $g(\varphi)$ according to (1) and (4). Dashed lines at $\pm\theta_{\text{db}}$ delimit the null-assistance band.

### A. Phase-Dependent Impedance Landscape

The online estimation of the gait phase $\varphi \in [0, 2\pi]$ relies on a pool of adaptive frequency oscillators [41], [42] that take the measured trajectory $\theta_m$ as input. The smooth phase error compensator introduced in [43], is used to align the null values of $\varphi$ with the heel-strike (HS) events. This gait phase estimation method is commonly used in oscillator-based controllers for active orthoses and exoskeletons [44], therefore it is not described here for the sake of brevity.

The estimated gait phase $\varphi$ acts as the pace-maker for the $PI^2$-AAN controller. To this end, the gait cycle is divided into $P$ equal gait phase segments whose midpoints are indicated as $\varphi_1, \varphi_2, ..., \varphi_P$. A set of $P$ Gaussian basis functions centered at each $\varphi_i$ (i.e., $\Psi(\varphi) = \{\psi_1(\varphi), \psi_2(\varphi), ..., \psi_P(\varphi)\}^T$), and an associated set of shape parameters $\mathbf{w} = \{w_1, w_2, ..., w_P\}^T$ are used to form a phase-locked impedance landscape $g(\varphi)$, as depicted in Fig. 2(a). The magnitude of $g$ at a generic gait phase instant $\varphi$ is computed as

$$g(\varphi) = \frac{\sum_{i=1}^{P} \psi_i(\varphi) w_i}{\sum_{i=1}^{P} \psi_i(\varphi)}, \quad (4)$$

where $\psi_i(\varphi)$ indicates the Gaussian basis function centered at $\varphi_i$ and evaluated at $\varphi$

$$\psi_i(\varphi) = e^{-0.5\,\mu\,(\varphi - \varphi_i)^2}, \quad (5)$$

and the constant parameter $\mu$ defines the width of the basis functions.

The gait-phase parameterization decouples the basis functions from the lapsed time and ensures consistent indexing of the impedance landscape (4) within each gait cycle, despite stride-to-stride variations in the duration of each cycle, which typically occur during walking. Furthermore, owning to the vast expressive power of the basis functions [45], smooth trajectories of arbitrary shape can be created by a proper set of shape parameters $\mathbf{w}$ [36], [46]. An example of force field resulting from (1) and (4) is illustrated in Fig. 2(b).

### B. On-line Adaptation of Impedance Landscape Using $PI^2$

The $PI^2$ algorithm approaches the reinforcement learning problem from the viewpoint of nonlinear stochastic optimal control and learns optimal policies through Monte-Carlo evaluations [36]. $PI^2$ generalizes the path integral control framework [47], [48] and extends its range of application to stochastic dynamic systems. In past studies, $PI^2$ has been successfully applied to high dimensional planning and control problems [49], [50]. The core of the $PI^2$ algorithm consists of a *parameterized policy* that is represented by a set of shape parameters $\mathbf{w}$. Policy improvement is achieved iteratively through a repeating sequence of *system trajectory sampling*, *policy update*, and *policy evaluation*. This sequence forms an *epoch*. Within each epoch, $K$ exploration roll-outs with stochastic actions $\mathbf{w} + \boldsymbol{\epsilon}$ (with $\boldsymbol{\epsilon}$ indicating the exploration noise) are first sampled, followed by a parameter update that assigns higher weights to the actions that demonstrate higher probabilities of lowering the overall trajectory cost in the sampled roll-outs. Immediately after the parameter update, a noiseless roll-out with the updated shape parameters $\mathbf{w}$ is carried out to evaluate the new policy. Subsequently, the trajectory cost measured in the evaluation trial serves as a performance metric that determines the next action: terminate the policy improvement process if a specified convergence criterion is met, or proceed to the next epoch otherwise.

To tailor the original $PI^2$ framework for RAGT applications we treat each gait cycle (stride) as one roll-out and regard the shape parameters $\mathbf{w}$ defining the impedance landscape (4) as the parameterized policy to be optimized. Additionally, the discrete-time learning procedure of $PI^2$ is replaced with a gait phase-indexed approach that updates the $PI^2$ trajectory costs at $N$ equally spaced gait phase instants $\varphi_1, \varphi_2, ..., \varphi_N$. Similar to the phase-dependent impedance landscape described in Sec. II-A, a gait phase segment of width $\frac{2\pi}{N}$ is associated with each phase instant. The update procedure of the modified $PI^2$ can be summarized as follows:

i) For each stride $k$ in the current set of $K$ exploration roll-outs, compute $S(\xi_{n,k})$, the cost-to-go associated with a partial trajectory $\xi_{n,k}$ that starts at $\varphi_n$ and extends to the end of the stride. This is achieved using forward integration:

$$S(\xi_{n,k}) = \sum_{j=n}^{N} r_{j,k} + \frac{1}{2} \sum_{j=n}^{N} \mathbf{W}_{j,k}^T \mathbf{R} \, \mathbf{W}_{j,k} \quad (6a)$$

for all $n = 1, ..., N$ and $k = 1, ..., K$, where

$$\mathbf{W}_{j,k} = \mathbf{w} + \mathbf{M}_j \, \boldsymbol{\epsilon}_{j,k} \quad (6b)$$

$$\mathbf{M}_j = \frac{\mathbf{R}^{-1} \, \boldsymbol{\Psi}(\varphi_j) \, \boldsymbol{\Psi}(\varphi_j)^T}{\boldsymbol{\Psi}(\varphi_j)^T \, \mathbf{R}^{-1} \, \boldsymbol{\Psi}(\varphi_j)}. \quad (6c)$$

$\mathbf{M}_j \in \mathbb{R}^{P \times P}$ is a projection matrix designed to eliminate the policy parameters that do not contribute to the overall control cost. It depends on $\boldsymbol{\Psi}(\varphi_j) \in \mathbb{R}^P$, the vector of $P$ Gaussian basis functions (5) evaluated at phase instant $\varphi_j$, and $\mathbf{R} \in \mathbb{R}^{P \times P}$, a positive semi-definite matrix that weighs the control cost. $\boldsymbol{\epsilon}_{j,k} \in \mathbb{R}^P$ is the exploration noise drawn at stride $k$ and phase $\varphi_j$ from a distribution $\mathcal{N}(0, \sigma^2)$. $r_{j,k}$ indicates the immediate cost of the current policy computed at stride $k$ across the $j$-th gait phase segment centered at $\varphi_j$:

$$r_{j,k} = \lambda_\theta \, \overline{\Delta \theta}_{j,k}^2 + \lambda_g \, g^2(\varphi_j). \quad (7)$$

$\overline{\Delta \theta}_{j,k}$ is the root-mean-square (RMS) tracking error across the $j$-th phase segment and $g(\varphi_j)$ is the impedance landscape (4) evaluated at the center of the same segment. Thus, $r_{j,k}$ is a weighted sum of movement errors and robot assistance accrued over the phase segment associated with $\varphi_j$. The weights $\lambda_\theta$ and $\lambda_g$ are constant within the $K$ exploration strides, however they are updated every $M$ epochs, using the high-level policy evaluation rules described in Sec. II-C. Equation (6a) is evaluated for all $N$ gait-phase instants, across all the $K$ exploration strides, thereby resulting in $K \times N$ estimations of the cost-to-go $S(\xi_{n,k})$.

ii) Compute the discrete probability associated with each cost-to-go

$$\mathbb{P}(\xi_{n,k}) = \frac{e^{-\frac{1}{\lambda} S(\xi_{n,k})}}{\sum_{k=1}^{K} [e^{-\frac{1}{\lambda} S(\xi_{n,k})}]}, \quad (8)$$

such that partial trajectories resulting in lower costs $S(\xi_{n,k})$ are designated with a higher value of $\mathbb{P}(\xi_{n,k})$. The sensitivity of $\mathbb{P}(\xi_{n,k})$ to each exponentiated $S(\xi_{n,k})$ is modulated by $\lambda$. As suggested in [36], one can optimize $\lambda$ for every phase instant $\varphi_n$, $n = 1, ..., N$, to maximally discriminate among the $K$ exploration roll-outs. This is equivalent to replacing the exponential terms in (8) with

$$e^{-\frac{1}{\lambda} S(\xi_{n,k})} = e^{-h \frac{S(\xi_{n,k}) - \min_j S(\xi_{n,j})}{\max_j S(\xi_{n,j}) - \min_j S(\xi_{n,j})}}, \quad (9)$$

where $h$ is a user-defined constant, and the functions min and max are taken over all $K$ exploration strides [36].

iii) Compute the update $\delta \mathbf{w} = \{\delta \mathbf{w}_1, \delta \mathbf{w}_2, ..., \delta \mathbf{w}_P\}^T$ by first taking the probability-weighted average at each phase instant $\varphi_n$, along the direction of the $K$ explored strides

$$\delta \mathbf{w}_n = \sum_{k=1}^{K} \mathbb{P}(\xi_{n,k}) \, \mathbf{M}_n \, \boldsymbol{\epsilon}_{n,k} \quad n = 1, ..., N, \quad (10)$$

and then by taking a second average along the direction of the $N$ phase instants:

$$\delta \mathrm{w}_i = \frac{\sum_{n=1}^{N} (N-n) \, \psi_i(\varphi_n) \, \langle \delta \mathbf{w}_n, \mathbf{e}_i \rangle}{\sum_{n=1}^{N} (N-n) \, \psi_i(\varphi_n)} \quad i = 1, ..., P. \quad (11)$$

In the previous expression, $\psi_i(\varphi_n)$ indicates the $i$-th basis function as defined in (5), while $\langle \delta \mathbf{w}_n, \mathbf{e}_i \rangle$ is a scalar product yielding the $i$-th entry of vector $\delta \mathbf{w}_n \in \mathbb{R}^N$ defined in (10). By evaluating (11) for all $P$ indices, we obtain the update $\delta \mathbf{w} \in \mathbb{R}^P$. From (8) and (10), it can be inferred that $\mathbb{P}(\xi_{n,k})$ determines the weighting of local controls with the cost $S(\xi_{n,k})$, such that the contributions of the explorations that resulted in lower costs are maximized.

iv) Update the policy parameters using $\delta \mathbf{w}$:

$$\mathbf{w} \leftarrow \mathbf{w} + \delta \mathbf{w}. \quad (12)$$

It is worth noting that the number of gait phase instants $N$ determines the granularity with which the exploration roll-outs are evaluated, whereas the number of basis functions $P$ defines the granularity of the parameterized policy. Thus, $N$ and $P$ are independent parameters.

The PI$^2$ algorithm adjusts the policy parameters directly via a learning rule, which optimizes over states-action trajectories $\xi_k = (\varphi_1, g(\varphi_1), ..., \varphi_N, g(\varphi_N))$, without the need for a value function to be computed explicitly [50]. This type of direct policy learning from sampled roll-outs (strides) lends itself to applications wherein prior knowledge of the environment is limited. Thus, it is particularly suitable for robot-assisted rehabilitation, in which an accurate model of the human-robot co-adaptation processes is typically very difficult to obtain [13]. Furthermore, the absence of matrix inversions and gradient learning rates in the update equations of PI$^2$ allows for better computational efficiency and numerical robustness compared to other probabilistic algorithms and policy gradient algorithms [36], eluding numerical pitfalls that can potentially be induced by the varying human motor capabilities [37].

### C. Hierarchical Policy Parameter Evaluation

The processes of human motor learning and human-robot co-adaptation involve complex dynamics that can hardly be represented by a single objective function. In traditional therapist-assisted gait training, the level of physical assistance changes depending on the subjects' performances. Assistance can be substantial at times, to support intense exercise regimens for severely impaired subjects [51], but it may be progressively reduced over time to keep the subjects challenged throughout the training sessions [52]. From the perspective of the learning algorithm, this translates into the need for a varying cost function that can specify different learning goals based on the subject's training progress. To this end, we propose a dual-objective convergence rule that yields large assistance when subjects deviate too far away from the target path (*intervention* mode), and relaxes the assistance if the tracking performance is satisfactory over a certain time period (*compliance* mode). The two modes share the same form of the immediate cost (7), however, the weight vector $\mathbf{\Lambda} = \{\lambda_\theta, \lambda_g\}^T$ is configured as $\lambda_\theta \gg \lambda_g$ in *intervention* mode to prioritize tracking accuracy, and as $\lambda_\theta \ll \lambda_g$ in *compliance* mode to facilitate the gradual decrease of robotic assistance and thus encourage subject's participation. The alternation between the two learning modes is determined by a hierarchical evaluation structure, as shown in Fig. 3.

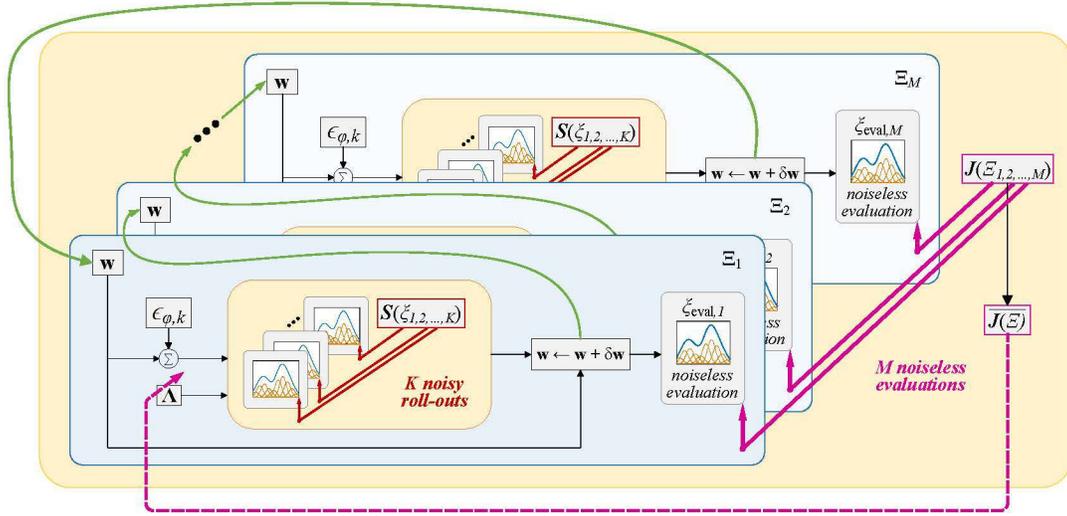

Fig. 3. Hierarchical evaluation structure of the PI$^2$-AAN controller: $K$ exploration roll-outs (strides) carried out at the lower level are used to update the policy parameters $\mathbf{w}$; $M$ high-level costs $J(\Xi_m)$ accumulated over noiseless evaluations are used to determine the learning mode (i.e., *intervention* or *compliance*) and the corresponding weight vector $\mathbf{\Lambda} = \{\lambda_\theta, \lambda_g\}^T$ for the next iteration.

At the low level, trajectory costs $S(\xi_{n,k})$ collected during $K$ exploration strides within the current epoch $\Xi_m$ are used to update the policy parameters $\mathbf{w}$, as described in Sec. II-B. Subsequently, a noiseless evaluation stride $\xi_{\text{eval},m}$ is carried out to determine the cost $J(\Xi_m)$ associated with the current epoch, which is quantified as the RMS tracking error accrued over the evaluation stride:

$$J(\Xi_m) = \overline{\Delta\theta}(\xi_{\text{eval},m}) \qquad (13)$$

In contrast to the conventional PI$^2$ algorithm, convergence of the current learning mode is not based solely on the cost $J(\Xi_m)$. Instead, we regard $J(\Xi_m)$ as a temporal abstraction of the cost associated with $\Xi_m$, and evaluate the current policy at a higher level, from the costs collected across $M$ consecutive epochs: $J(\Xi_1), J(\Xi_2), ..., J(\Xi_M)$. The average of these higher level costs, denoted as $\overline{J(\Xi)}$, determines the termination/switching condition specified by the current learning mode:

$$\begin{cases} \textit{compliance} \to \textit{intervention} \text{ mode,} & \text{if } \overline{J(\Xi)} > \beta^u, \\ \textit{intervention} \to \textit{compliance} \text{ mode,} & \text{if } \overline{J(\Xi)} < \beta^l. \end{cases} \qquad (14)$$

The upper error bound $\beta^u$ represents the largest cumulative errors one is allowed to make while the orthosis is controlled in *compliant* mode, above which the *intervention* mode will be activated to guide the limb motion towards the desired pattern with stiffer assistance. Conversely, $\beta_l$ represents the smallest cumulative errors for which the subject receives stiff assistance from the orthosis, before the *compliance* mode takes over and begins reducing the assistance level to discourage the subject from over-relying on the robot assistance. After the onset of the *compliance* mode, two alternative scenarios may occur: i) the subject has *not learned* the desired motion, which results in an increase in tracking errors that causes $\overline{J(\Xi)}$ to eventually reach $\beta^u$ and trigger the *intervention* mode; or ii) the subject has *learned* the motion and is able to maintain the tracking error below the acceptable error bound $\beta^u$, even when the robotic assistance has diminished to nearly zero. The latter scenario, in which the algorithm appears to be "trapped" in the *compliance* mode, is regarded as an indicator of human motor adaptation to the target trajectory.

To allow the human-robot co-adaptation process to eventually reach a steady-state, a decay parameter $\gamma$ is imposed on $\sigma^2$ (the variance of the exploration noise $\epsilon$), which has the effect of mitigating the changes to the adaptive force field over time. Sufficient exploration is ensured by resetting $\sigma$ to its starting value each time the learning mode is changed.

A brief overview of the update procedure of the PI$^2$-AAN algorithm is listed in Algorithm 1, while the source code is made available as supplementary material[1]. It is worth noting that, since the high-level cost (13) is determined exclusively by the tracking errors, the level of robotic assistance accounted for in the lower level cost (7) is disregarded at the higher level when computing the switching conditions (14). Therefore, the PI$^2$-AAN algorithm can operate indefinitely during each RAGT session to continuously optimize the assistance level with the goal of facilitating the human motor adaptation process. This allows the PI$^2$-AAN controller to mimic the behavior of a physical therapist who constantly modulates the

---

[1] https://github.com/wearable-robotic-systems-lab/Powered-Orthosis

**Algorithm 1** Pseudocode of the PI$^2$-AAN algorithm

*Input*:
- $P$: number of Gaussian kernel functions forming the parameterized policy.
- $N$: number of discrete phase instants for exploration roll-outs.
- $K$: number of exploration strides performed before policy updates.
- $M$: number of epochs performed before high-level evaluations.
- $\mu$: width of each kernel function.
- $\sigma^2, \gamma$: variance and decay parameter determining the distribution $\mathcal{N}(0, \gamma^{(\text{\# of updates since last reset})} \sigma^2)$, from which the exploration noise $\epsilon$ is drawn.
- $\mathbf{\Lambda}$: learning mode weight vector.
- $\beta^u, \beta^l$: error bounds determining the learning mode.

*Initialize*:
- $\mathbf{w}_{\text{init}}$: initial policy parameters
- $J_{\text{init}}$: initial high-level cost for learning mode selection

**while** RAGT session is *in progress* **do**
    Compute $\overline{J(\Xi)}$ and determine learning mode using (14)
    Reset $\gamma$ if selected mode differs from previous iteration
    Carry out $M$ epochs to evaluate high-level cost $J$:
    **for** $m = 1, 2, ...M$ **do**
        Perform $K$ exploration roll-outs to update $\mathbf{w}$:
        **for** $k = 1, 2, ...K$ **do**
            **for** $j = 1, 2, ...N$ **do**
                Evaluate $r_{j,k}$ using (7)
                Compute $\mathbf{W}_{j,k}, \mathbf{M}_j$ using (6b), (6c)
            **end for**
            Evaluate low-level cost-to-go $S(\xi_{n,k})$ using (6a), (7)
        **end for**
        Compute $\mathbb{P}(\xi_{n,k})$ using (8), (9)
        Update $\mathbf{w}$ using (10), (11), (12)
        Deploy noiseless $g(\varphi)$ for one stride using (4)
        Compute $J(\Xi_m)$ using (13)
    **end for**
**end while**

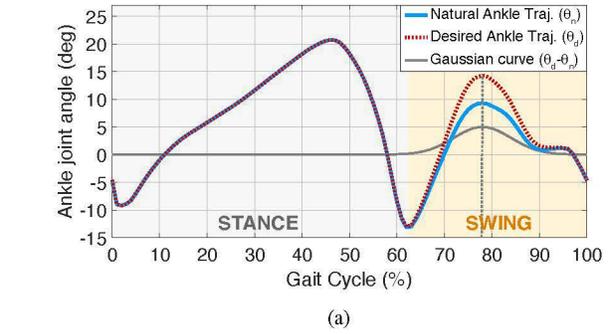

(a)

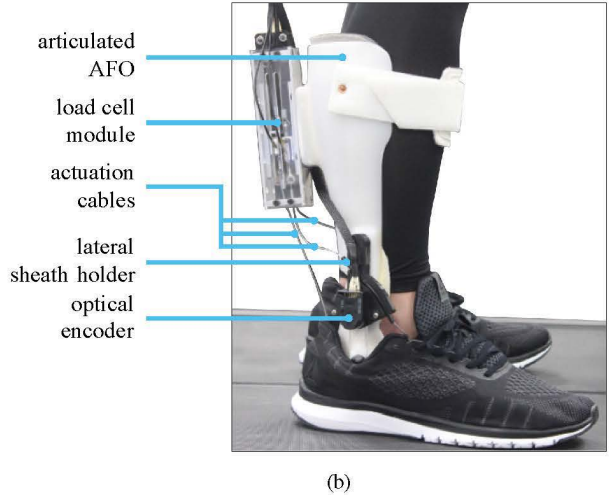

(b)

Fig. 4. (a) Average ankle trajectory measured from a representative subject during treadmill stepping under null orthosis assistance (blue line). The desired trajectory $\theta_d$ (orange dashed line) is defined as the sum of the subject's average trajectory with no assistance and a Gaussian curve (gray solid line). The vertical dashed line indicates the center of the Gaussian curve, which is located at the gait phase where the ankle joint reaches the maximum dorsiflexion angle during swing. (b) On-board components of the SAFE orthosis.

assistance level while pursuing the long-term goal of coaching patients to perform the desired motion by themselves.

### III. EXPERIMENTS

To validate the PI$^2$-AAN algorithm in RAGT, walking experiments involving learning of a new ankle gait pattern were carried out with a group of able-bodied individuals. A powered ankle-foot orthosis was used to provide assistance to the study participants during training.

#### A. Motor Learning Task

Sufficient foot clearance is closely related to an individual's ability to dorsiflex the foot during the swing phase and is a key element of stable ambulation [53]. The lack of ankle dorsiflexion during swing is a common impairment among stroke survivors due to overreactivity of plantarflexor muscles, inability to generate sufficient dorsiflexion moment, or shortening of the plantar flexor muscles [54]. Recent studies on RAGT have shown that the foot clearance can be efficiently increased by supplying upward assistive forces to subject's ankle via an impedance-controlled exoskeleton [55]. Studies have also found that the minimum toe clearance, which typically takes place at approximately 51% of the swing phase, plays a critical role in foot clearance and can be increased by enlarging the ankle dorsiflexion angle at that phase instant [53], [56].

In this vein, we specify the RAGT task as an ankle movement that amplifies one's ankle dorsiflexion angle during midswing. As shown in Fig. 4(a), the desired trajectory $\theta_d$ is defined as the sum of the subject's natural gait pattern and a Gaussian curve centered at the gait phase where his/her ankle joint reaches the maximum dorsiflexion during swing. The magnitude of the Gaussian curve is set to 5 $deg$ for all participants, and its width is configured to fade the altered trajectory to effectively zero in the stance phase. Since the normal ankle range of motion in the dorsiflexion direction during locomotion varies between approximately 5 and 20 $deg$ [57], an additional 5 $deg$ of dorsiflexion represents a marked modification to an individual's baseline gait. Moreover, this approach proved to be an effective means to induce measurable changes to a subject's natural gait while not endangering gait stability [33].

#### B. Apparatus

The Stevens Ankle-Foot Electromechanical (SAFE) orthosis, shown in Fig. 4(b), is a cable-driven device built off of a modified articulated ankle-foot orthosis. An antagonistic mechanism formed by a pair of Bowden cables allows the device to provide active torque control over plantarflexion and dorsiflexion. Two BLDC motors (EC45, Maxon, Switzerland) placed on an off-board platform control the tension in each Bowden-cable. A load cell (LSB200, Futek, USA) is connected in-line with each cable to close the inner torque control loop (Fig. 1). A quadrature encoder is used to measure the ankle joint angle and close the outer position control loop of the impedance controller. Two force sensitive resistors are located underneath the calcaneus and the hallux to detect HS and toe-off events, respectively. Data acquisition and high/low-level control are carried out by a myRIO board (National Instrument, USA) with the low-level torque control loop running at 400 $Hz$ and the high-level RL-AAN controller at 100 $Hz$. The total weight of the worn components of the SAFE orthosis is 1.04 $kg$. More details on the device can be found in [58].

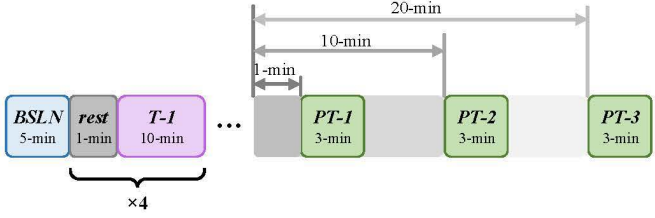

Fig. 5. Experimental protocol (*BSLN*: baseline session; *T-1~4*: training sessions; *PT-1~3*: post-training sessions).

### C. Experimental Protocol

A total of 10 able-bodied subjects (all males, age 28.4±0.5 years, weight 72.5±6.4 kg, height 176.2±2.9 cm) participated in the treadmill-walking tests. Participants were selected such that their right foot and lower leg could fit comfortably in the SAFE orthosis. The study was approved by Stevens Institutional Review Board and all participants provided informed consent prior to testing.

The experimental protocol is illustrated in Fig. 5. A 5-min walking bout was first administered to evaluate the subjects' baseline (*BSLN*) gait. During the *BSLN* session, the orthosis was controlled in transparent (i.e., zero-impedance) mode [59]. The average ankle trajectory recorded from the last minute of this session was regarded as each subject's natural gait pattern and was used to derive his target trajectory, as described in Sec. III-A. Afterwards, subjects underwent four 10-min training sessions (*T-1~4*) to learn the target ankle motion with the assistance provided by the SAFE orthosis, which was then set to operate under the PI$^2$-AAN controller. Four 1-min breaks were given before and after each *T* session. Following the *T* sessions, three post-training sessions (*PT-1~3*) were carried out, in which the SAFE orthosis operated under transparent mode, to examine training effects. The *PT* sessions started 1 minute, 10 minutes, and 20 minutes after the end of the last *T* session, respectively. The treadmill was paced at 1 $m/s$ for all walking bouts. The first 30 seconds of each walking bout were allocated as a ramp-up period to allow the treadmill speed, the subject's walking pattern, and the gait phase estimator to reach steady state.

### D. Parameter Selection

The PI$^2$-AAN parameters were selected as follows:

- $P = 10$, $\mu = 5$. The selected $P$ exceeds the number of functional sub-phases of the gait cycle [57], and therefore it was deemed appropriate to match different shape parameters with different functional gait tasks. $\mu$ was tuned to provide sufficient overlapping between the Gaussian kernels given the selected $P$.
- $N = 10$, $K = 4$, $M = 4$. $N$ was chosen as a trade-off between granularity in the evaluation of the exploration roll-outs and computational complexity of the controller. By performing $K = 4$ exploration roll-outs before each evaluation trial and $M = 4$ epochs before each high-level evaluation, the subject's performance and the effectiveness of the current learning mode were evaluated every 20 strides.
- $\sigma = 0.03$, $\gamma = 0.992$. $\sigma$ was selected to ensure sufficient exploration while preventing abrupt stride-to-stride changes in the impedance landscape during the exploration roll-outs. $\gamma$ was chosen to ensure that the algorithm continued to explore in the event that the same learning mode persisted for an extended period of time. Quantitatively, the selected decay factor reduces $\sigma$ to approximately 20% of its initial value after 200 strides.
- $\beta^u = 1.5$ $deg$, $\beta^l = 0.5$ $deg$. Because the target motor task involved alteration to the swing-phase ankle trajectory, we limited the performance evaluation (13) to the RMS tracking error measured during the latter half of the gait cycle (i.e., $\varphi_{6\sim10}$). The average cost $\overline{J(\Xi)}$ was then compared with the error bounds $\beta^{u,l}$ to determine the learning mode for the next iteration.
- $\mathbf{\Lambda} = \{80, 5\}^T$ and $\{5, 80\}^T$ in *intervention* and *compliance* mode, respectively.
- $\mathbf{w}_{\text{init}} = 0$, $\mathbf{J}_{\text{init}} = 2.5$ $deg$. All $T$ sessions started with a "flat" impedance landscape and with the *intervention* mode, to gradually conform the landscape to the subject's performance.
- $\tau_{\max} = 5$ $Nm$, $\theta_{\text{db}} = 1$ $deg$. These values were determined based on previous work [33], so that the assistive torque (1) could be modulated from effectively zero to a sufficiently large value to guide the subject's ankle joint.

### E. Data Analysis

To investigate subjects' motor adaptation processes, we analyzed steady-state ankle trajectories measured during the last 9 minutes of the *T* sessions and the last minute of the *PT* sessions. HS events were used to segment these trajectories into gait cycles, and RMS errors between the target and the segmented measured trajectories were used as the main performance metric indicating gait adaptation. To this end, one sample t-tests with Bonferroni correction were carried out on the RMS errors to check for significant differences ($\alpha = 0.05$) between subjects' performances in *BSLN* and *PT* sessions. Moreover, the phase-dependent impedance (4) evaluated at the kernel centers (i.e., $g(\varphi_1)$, ..., $g(\varphi_2)$, ..., $g(\varphi_{10})$) was logged at each gait cycle during the *T* sessions, to investigate the effectiveness of the PI$^2$-AAN algorithm in shaping individualized impedance landscapes. A Boolean variable indicating the *on/off* state of the *intervention* mode was also recorded to provide additional insights into the human-robot co-adaptation processes. Linear regression was performed on the group averages of each $g(\varphi_i)$, as well as on the percentage *on*-time of the *intervention* mode over each $T$ session, to examine the changes of subjects' reliance on the robot assistance. All data analysis was conducted using custom scripts developed in MATLAB (MathWorks, USA).

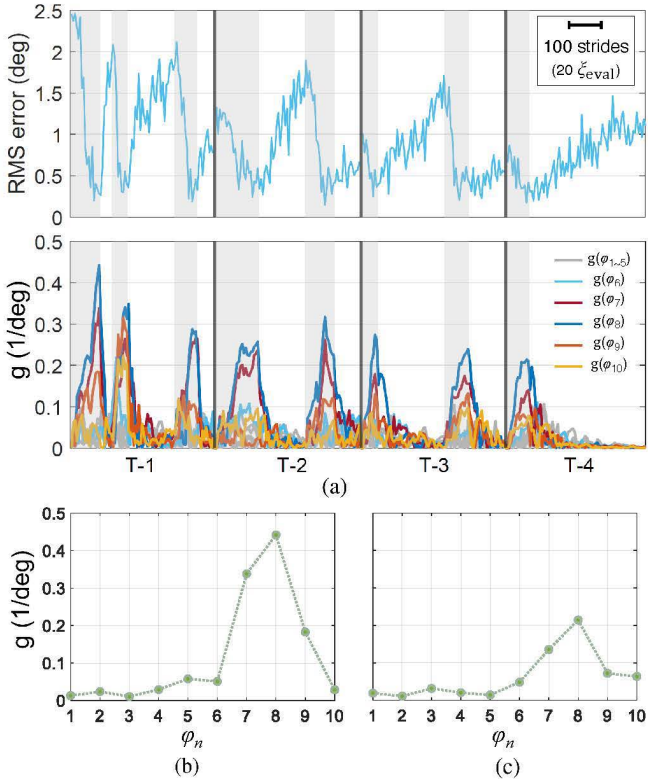

Fig. 6. (a) Change of RMS errors $J(\Xi_m)$ (top) and impedance coefficients $g(\varphi_i)$ (bottom) extracted from the evaluation strides $\xi_{\text{eval},m}$ occurring during the $T$ sessions of a representative subject. Each vertical line delimits a 10-minute training bout, while shaded areas indicate the periods in which the *intervention* mode was engaged. Swing-phase impedance coefficients $g(\varphi_6) \sim g(\varphi_{10})$ are color-coded to evidence their larger magnitudes compared to their stance-phase counterparts $g(\varphi_1) \sim g(\varphi_5)$. The stiffest impedance landscapes measured in *T-1* and *T-4* from the same subject are shown in (b) and (c), respectively.

## IV. Results

Figure 6(a) shows the training progress of a representative subject and the trend of the learned impedance coefficients measured in the evaluation strides throughout the $T$ sessions. The shaded/non-shaded areas correspond to the periods in which the *intervention*/*compliance* modes were engaged, respectively. Noticeably, the *intervention* mode was effective in increasing the impedance magnitudes (and thereby the assistance levels) to help the subject follow the target movement. This scenario was consistent across the four training sessions. When the average RMS error $\overline{J(\Xi)}$ fell below $\beta^l = 0.5\ deg$, the *compliance* mode came into effect and began to progressively reduce the assistance levels to encourage the subject's active participation, as can be observed from the swiftly decreased impedance magnitudes immediately following the onset of each *compliance* period. The rate at which the tracking accuracy deteriorates after the onset of the *compliance* mode affects its duration. Therefore, inspecting the duration of the *compliance* mode across the $T$ sessions provides insights into the subject's motor adaptation progress. For example, the subject's insufficient ability to follow the target motion during the first *compliance* period in *T-1* resulted in a rapid increase in tracking errors that quickly triggered a switch-back to the *intervention* mode. In contrast, the *compliance* mode remained in effect for most of the *T-4* session as the subject maintained the average tracking error below $\beta^u = 1.5\ deg$, even after the assistance had faded to nearly zero. Because the target movement remained the same throughout the training sessions, the magnitudes of the impedance coefficients $g(\varphi_i)$ during the *intervention* mode can be interpreted as a surrogate measure of the amount of external assistance the subject required in order to follow the target movement. Thus, the overall downward trend of the impedance magnitudes from *T-1* to *T-4* suggests an increase in the subject's participation and an improvement in his ability to follow the target movement. Figure 6(b) and (c) show the impedance coefficients $g(\varphi_1)$ to $g(\varphi_{10})$ occurring in *T-1* and *T-4* at the gait cycle wherein the impedance landscape was the stiffest for the session. Evidently, the stiffest force field generated in *T-4* was less pronounced compared to the one in *T-1*. Since the intended training was designed to only alter subject's gait in the swing phase, the relatively small magnitudes of $g(\varphi_1) \sim g(\varphi_5)$ further indicate the ability of the PI$^2$-AAN algorithm to generate assistance only at the gait phases where it is needed. Moreover, the magnitude of $g(\varphi_8)$ conforms to the training task, which imposed the largest alternation to the subject's natural motion at approximately 80% of the gait cycle (Fig. 4(a)).

Group-wise averages of the impedance coefficients $g(\varphi_6) \sim g(\varphi_{10})$ during the *intervention* periods are shown in Fig. 7(a), alongside the percentage *on*-time of the *intervention* mode. Consistent with observations from the representative subject in Fig. 6(a), we notice that the largest impedance coefficient was $g(\varphi_8)$, while the other coefficients quickly diminished as their associated phases move away from $\varphi_8$. This suggests that the resulting impedance landscapes adapted well to the gait modification pattern shown in Fig. 4(a). The linear regression analysis shown in Fig. 7(a) evidenced a significant decreasing trend in both impedance coefficients and percentage *on*-time of the *intervention* mode over the four $T$ sessions, indicating that the participants were able to progressively adapt to the target trajectory and reduce their reliance on the robot assistance. The decreasing trends also highlight the effectiveness of the hierarchical evaluation structure in specifying an appropriate learning mode that fades assistance when subjects are performing well. In this respect, it is worth noting that, unlike $g(\varphi_7)$ and $g(\varphi_8)$, the other impedance coefficients did not strictly follow a downward trend and were instead kept at relatively small magnitudes throughout the training. This can be explained by the design of the cost function in the *intervention* mode, which moderately penalizes the control effort that does not significantly contribute to the reduction of tracking errors. In other words, the prioritization of $g(\varphi_7)$ and $g(\varphi_8)$ indicates that the robot assistance during 70~80% of the gait cycle played the most crucial role in helping individuals achieve the training task.

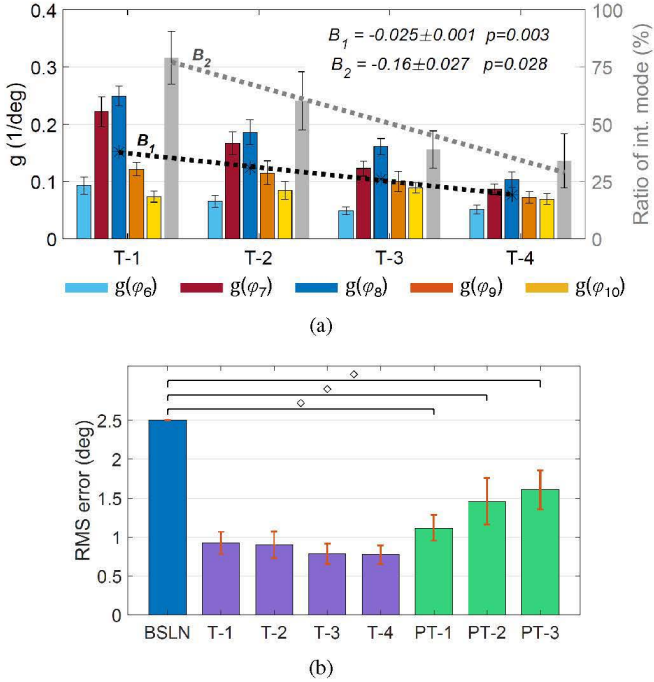

Fig. 7. (a) Group-wise average magnitudes of the impedance coefficients $g(\varphi_6) \sim g(\varphi_{10})$ during the *intervention* periods in the four $T$ sessions. The gray bars indicate the percentage *on*-time of the *intervention* mode within each $T$ session. $B_{1,2}$ are the angular coefficients of the regression lines obtained from the within-session averages of $g(\varphi_6) \sim g(\varphi_{10})$ and from the percentage *on*-time of the *intervention* mode across *T-1~4*, respectively. (b) Group averages of the RMS errors across all walking bouts. ⋄ indicates $p < 0.0001$ (Bonferroni-adjusted $p$ values). Error bars indicate ±1SE (standard error).

Figure 7(b) shows group averages of the RMS errors in all walking bouts. Because the target motion pattern is derived from each individual's baseline gait with the addition of the same Gaussian curve, the baseline RMS error is the same (i.e., approximately 2.5 $deg$) for all subjects. A slight decrease in average tracking errors can be seen in *T-3,4* compared to *T-1,2*, which, along with the decreased magnitude and duration of the assistance discussed above, suggest an overall improvement of subjects' ability to follow the target motion. Additionally, the one sample t-tests revealed significant retention of the target movement across *PT-1~3*, reflecting the efficacy of the proposed control strategy in promoting short-term motor adaptations.

## V. Discussion

Control strategies that promote subjects' active participation during robot-assisted training have been widely accepted as effective means to facilitate motor recovery [8]–[10]. In particular, impedance control-based assistive strategies have gained significant interests due to their compliant nature, allowing for small kinematic errors that are critical enablers of motor (re)learning [3], [13], [60]. To maintain a proper balance between kinematic errors and subjects' physical efforts, adaptive AAN controllers such as the ILC-based AAN controllers were introduced in pioneering works [11], [12]. These ILC schemes, however, typically require tedious manual tuning. This paper introduced a new AAN control approach for lower-extremity robotic trainers based on the RL paradigm, which allows an underlying impedance controller to self-modulate the control stiffness in a phase-dependent manner given a subject's motor capability and training progress. Unlike recent Bayesian optimization-based AAN methods [61], the proposed controller relies on the amount of assistance provided by the robotic trainer as a proxy to gauge subjects' active efforts, without the need for surface EMG measurements. The phase-dependent adaptability was made possible through the parameterization of the impedance landscape via phase-locked basis functions and through the use of a modified $PI^2$ algorithm. The correlation between a subject's training performance and the stiffness of the force field was established through a hierarchical evaluation structure that continuously alters the controller's priority between tracking accuracy and control effort, based on the subject's recent motor behaviors. Since the robot's control effort complements the subject's physical effort in a given training task, the proposed $PI^2$-AAN algorithm implicitly took subjects' effort into account when regulating the assistance levels.

Our previous works on ADHDP-AAN controllers successfully leveraged the reinforcement learning paradigm to establish adaptive correlations between the control stiffness of a powered orthosis and the subject's training performance [29], [33]. However, in those ADHDP-AAN controllers the granularity of the adaptation was limited to the stride level, whereby the same stiffness parameter was enforced across the entire gait cycle. The main contribution of this paper is the extension of the control adaptability to enable modulations at the phase level. To achieve this goal, we introduced a new $PI^2$-based strategy that explicitly learns and updates the impedance coefficient in a phase-dependent fashion. Although it is feasible to setup parallel ADHDP blocks for independent phases and enable a similar phase-dependent adaptability [24], the independent update of ADHDP blocks may lead to discontinuous inter-phase actions due to the lack of extrapolation across contiguous blocks. On the contrary, by parameterizing the impedance landscape through a vector of basis functions, each phase-dependent impedance value derived by the $PI^2$-AAN algorithm is jointly determined by a combination of weighted basis functions equally distributed along the gait phase axis, which intrinsically ensures smooth inter-phase transitions. Moreover, the probability-based learning of $PI^2$-AAN provided additional benefits that are well suited for RAGT. Firstly, instead of indiscriminately accepting all new sampled states to update the current knowledge about the environment (i.e., the human-robot interaction), the $PI^2$-AAN algorithm performs policy update by evaluating probability-weighted averages over multiple roll-outs (i.e., strides) and relies primarily on the ones that have higher probabilities of reducing trajectory costs, which greatly reduces the potential influence induced by the uncertain and varying human motor learning dynamics [62]. Secondly, the empirical evaluation of the sampled trajectories eliminates the need for calculating the gradient of a cost function, which represents a distinctive advantage over policy gradient methods like ADHDP for dealing with non-smooth cost functions constructed with noisy kinematic errors [20], [24], [33]. Thirdly, since the exploration of policy parameters is strictly controlled by the exploration noise $\epsilon$ and the updated policy parameters are confined within the convex hull encircled by $\mathbf{w}+\epsilon$ [50], [63], the control output is naturally bounded during the PI$^2$-AAN policy improvement process.

The proposed hierarchical approach to policy parameter evaluation stems from the hierarchical RL paradigm, in which temporally-extended activities are stimulated to benefit learning tasks that do not require immediate actions at each step [64]. Admittedly, the hierarchical structure proposed here does not assign separate sub-goals at different levels [63] but rather consolidates the optimality of the updated policy in the current mode by evaluating it over a longer term. In the context of RAGT, such temporally-extended evaluation is particularly valuable because it ensures that the optimized policy (i.e., the assistance level) can benefit the training subject in a consistent and repeatable manner [65]. Additionally, the hierarchical approach inherently prevents accidental convergence misguided by a single unstable stride.

The method introduced in this paper can be best compared to [55], [66], in which phase-dependent modulation of AAN controllers was achieved through the use of ILC-based update laws. The level of adaptability of these ILC-based approaches is primarily controlled by the so-called forgetting factor, which is typically determined based on simplistic assumptions about human motor learning dynamics [14]. Since the process of motor learning involves large uncertainties and the level of neurologic injury varies between patients, the applicability of such approaches to clinical populations may be limited [14], [37], [62]. In the PI$^2$-AAN framework, such limitation is circumvented by the dual-objective learning mode that automatically increases/decreases the assistance when a subject's recent performance deteriorates/improves. To this end, the subject's learning ability is not only evaluated online through their recent behaviors, but also directly linked to the learning goal of the PI$^2$-AAN algorithm that ultimately guides control decisions. Moreover, owing to the model-free nature of the PI$^2$, the parameter update procedure relies solely on probability-weighted sampled states rather than estimated dynamic models [67], [68] or approximated motor ability models [7]. As such, the proposed strategy may be more robust to variations in system dynamics or human motor learning abilities compared to model-based approaches. Besides, the known scalability of the PI$^2$ algorithm for high dimensional control tasks [69] makes this strategy applicable to most multi-DOF gait trainers that feature low-level force/torque control loops [16], [38], [55]. Thus, although the experimental validation performed in this work focused on gait training, the PI$^2$-AAN controller can potentially be extended to other repetitive rehabilitation tasks (e.g., arm reaching exercises for upper extremity rehabilitation [70]).

Future studies involving patients with neurological disorders must be carried out to investigate whether the PI$^2$-AAN controller's adaptability exhibited with able-bodied populations transfers to patients with gait disabilities, and to examine how patients would react to the continuously changing assistance levels. Moreover, a future study comparing the longitudinal effects of the PI$^2$-AAN strategy with those of the ADHDP-AAN controller and more conventional AAN controllers may provide additional insights into the benefits of individualization and adaptability in robot-assisted exercise therapy. The applicability of the proposed strategy to error-enhancing controllers, which leverage destabilizing force fields as opposed to stabilizing force fields, will also be examined [71]. Additionally, the robustness of the PI$^2$-AAN controller will be further evaluated through overground walking experiments. From an algorithm perspective, the adaptability of the PI$^2$-AAN controller may be extended by incorporating the Covariance Matrix Adaptation strategy to allow autonomous modulation of the exploration noise [72]. Further investigation and exploration on other types of kernel functions may also be valuable for improving the effectiveness and robustness of the proposed method. The magnitudes of the upper and lower error bound $\beta^{u,l}$ may also be adaptively regulated by specifying sub-goals at the higher level to progressively reduce the tolerable error region as subjects make progress, possibly through the addition of a performance-related terminal cost that penalizes the magnitudes of $\beta^{u,l}$ over time. Ongoing work also includes the study of exoskeleton design individualization as an additional means to personalize future robot-assisted rehabilitation interventions [73].

## VI. CONCLUSION

This work introduced a novel adaptive AAN controller for lower-extremity powered orthoses used in RAGT, which enables individualized and phase-dependent assistance. The proposed strategy consists of a modified PI$^2$ algorithm that shapes phase-locked impedance landscapes, and a temporally-extended evaluation structure that alters the PI$^2$ learning goal according to the subject's recent performances. Walking experiments conducted with a group of able-bodied individuals validated the effectiveness of the PI$^2$-AAN controller in providing individualized assistance and promoting short-term gait adaptations.

## ACKNOWLEDGMENT

This work was supported by the US National Science Foundation (Grant Number CMMI-1944203) and the New Jersey Health Foundation (Research Grants Program, Grant Number PC 6-18).